\pdfoutput=1
\documentclass[11pt]{article}
\usepackage{EACL2023}
\usepackage{times}
\usepackage{latexsym}
\usepackage[T1]{fontenc}
\usepackage[utf8]{inputenc}
\usepackage{microtype}
\usepackage{inconsolata}
\usepackage{hyperref}
\usepackage{xcolor}
\usepackage{graphicx, subcaption}
\usepackage{amsmath,amsfonts,amsthm,bm, amssymb, mathtools} 
\usepackage[shortlabels]{enumitem}
\usepackage{multirow,adjustbox,makecell,arydshln} 
\usepackage[ruled,linesnumbered]{algorithm2e}

\usepackage{pifont}

\renewcommand{\vec}[1]{\boldsymbol{\mathbf{#1}}}
\newcommand\Mycomb[2][^n]{\prescript{#1\mkern-0.5mu}{}C_{#2}}

\title{Do Neural Topic Models Really Need Dropout?\\Analysis of the Effect of Dropout in Topic Modeling}

\author{Suman Adhya,  Avishek Lahiri, Debarshi Kumar Sanyal \\
         Indian Association for the Cultivation of Science, Jadavpur, Kolkata-700032, India \\
         \texttt{\{ adhyasuman30, avisheklahiri2014, debarshisanyal \}@gmail.com}
}

\begin{document}
\maketitle
\begin{abstract}
Dropout is a widely used regularization trick to resolve the overfitting issue in large feedforward neural networks trained on a small dataset, which performs poorly on the held-out test subset. Although the effectiveness of this regularization trick has been extensively studied for convolutional neural networks, there is a lack of analysis of it for unsupervised models and in particular, VAE-based neural topic models. In this paper, we have analyzed the consequences of dropout in the encoder as well as in the decoder of the VAE architecture in three widely used neural topic models, namely, contextualized topic model (CTM), ProdLDA, and embedded topic model (ETM) using four publicly available datasets. We characterize the dropout effect on these models in terms of the quality and predictive performance of the generated topics.
\end{abstract}

\section{Introduction}
Dropout \cite{HintonDropout} is used while training neural networks, by stochastically dropping out the activation of neurons to prevent complex co-adaptations of feature vectors \cite{baldi_understanding_dropout}. The working of dropout is attributed to the implicit averaging over an ensemble of neural networks \cite{DropoutSurvey, DropoutWarde}. It has been shown to be effective on supervised learning tasks to prevent overfitting \cite{SrivastavaDropout}.

As the volume of digital documents significantly increases with time, organizing them manually is becoming quite an inconvenient task. Because of the ability of topic models to learn a thematic structure from a set of documents in an unsupervised manner and label the documents with their corresponding dominant topics, the significance of topic models is enormous in this area \cite{hall-etal-2008-studying, adhya-sanyal-2022-indian}. But in the traditional topic models, not only the computation cost of the approximate posterior is very high but also for a small change in the modeling assumption, re-derivation of the inference method is needed. With greater flexibility and scalability than traditional topic models, a class of Neural Topic Models (NTMs) aim to leverage the potential of neural networks using the AEVB \cite{VAE} based inference technique. Following \cite{NTMsurvey}, we refer to this class of models as VAE-NTMs where the training objective is to maximize the log-likelihood of the reconstruction of the input document while minimizing the KL-divergence of the learned posterior distribution of the latent space from a known prior distribution.

An earlier study by \cite{LDAdropout} of the dropout effect on two traditional topic models LDA \cite{LDA} and BTM \cite{BTM} shows that the correct choice of the dropout rate not only decreases the learning time of the models but also significantly improves the predictive performance and generalization for short texts. However, the study does not consider neural topic models.

In this work, we propose the use of dropout on VAE-NTMs as a hyperparameter in order to achieve much better performance in terms of topic coherence, topic diversity, and topic quality. We test this proposition on a range of standard VAE-NTM architectures.
To the best of our knowledge, there has been no other study focusing specifically on the use of dropout in neural topic models. We have made our analysis publicly available\footnote{\url{https://github.com/AdhyaSuman/NTMs_Dropout_Analysis}}.

In summary, our contributions are as follows: 
\begin{enumerate}
    \item We comprehensively show both quantitatively and qualitatively that topic quality undergoes a massive improvement with either very low or zero dropout settings in both the encoder and the decoder of a VAE-NTM.
    \item We show that for VAE-NTMs the systematic choice of low dropout rates can lead to a significant improvement in downstream tasks like document classification.
    \item We study the dependence of dropout on the length of the input documents.
    \item We present an empirical analysis for the increase in performance of VAE-NTMs with a decrease in dropout.
\end{enumerate}

\section{Task Formulation}
Given a corpus $\{D_1, D_2, \ldots, D_N\}$ of $N$ documents with vocabulary $\{w_1, w_2, ..., w_{V}\}$ of $V$ words, topic models describe a document $D_i$ as a distribution over $K$ topics $\{\vec{\beta}_1, \vec{\beta}_2, ..., \vec{\beta}_K\}$, where an individual topic $\vec{\beta}_k$ is a distribution over $V$-words.

\subsection{VAE Framework in Neural Topic Models}

\begin{figure}[ht]
    \centering 
    \includegraphics[width=.9\linewidth]{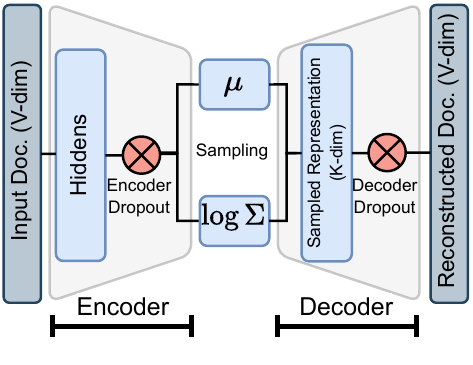}
    \caption{VAE framework in neural topic models.}
    \label{fig:VAE_Arch}
\end{figure}

Given an input sample $\vec{x}$, a VAE encoder learns the approximate posterior distribution $q_{W}(\vec{z}|\vec{x})$ where $W$ is the encoder's weights that are to be learned and $\vec{z}$ is a latent variable. Given a sample $\vec{z} \thicksim q_{W} (\vec{z}|\vec{x})$, the VAE decoder learns the likelihood $p_{W'}(\vec{x}|\vec{z})$ where $W'$ is the learnable decoder's weights.

In VAE-NTMs the input to the encoder is a document representation (e.g., bag-of-words) $\vec{x}_{V \times 1}$. The encoder then returns the Gaussian parameters $\left( \vec{\mu}_{K \times 1}, \vec{\Sigma}_{K \times 1} \right)$ that approximate the true posterior where $K$ is the dimension of latent (topic) space, $ \vec{\mu}_{K \times 1}$ is the mean, and $\vec{\Sigma}_{K \times 1}$ is the diagonal covariance matrix. 
Upon taking these Gaussian parameters as input, the decoder samples a latent representation $\vec{z}_{K \times 1}$ from $\mathcal{N}(\vec{\mu}_{K \times 1}, \vec{\Sigma}_{K \times 1})$ using the reparametrization trick as follows:
\begin{equation*}
    \vec{z}_{K \times 1} = \vec{\mu}_{K \times 1} + \vec{\Sigma}^{\frac{1}{2}}_{K \times 1} \odot \vec{\epsilon}_{K \times 1}
\end{equation*}
where $\vec{\epsilon}_{K \times 1} \thicksim \mathcal{N}(\vec{0}, \vec{I})$ and $\odot$ represents the element-wise product.
Then the document-topic distribution vector ($\vec{\theta}_{K \times 1}$) is generated such that $\vec{\theta}_{K \times 1} = \sigma( \vec{z}_{K \times 1} )$ where $\sigma(\cdot)$ is a softmax function.
The input document-term distribution vector is reconstructed with the product of $\vec{\theta}_{K \times 1}$ and $\vec{\beta}_{K \times V}$, the topic-word matrix, in the following manner:
\begin{equation*}
{\tilde{\vec{x}}}_{V \times 1} = 
\begin{cases}
 \vec{\beta}^T \vec{\theta} & \text{if } \vec{\beta} \text{ is normalized.} \\ 
 \sigma \left(\vec{\beta}^T \vec{\theta} \right) & \text{if } \vec{\beta} \text{ is unnormalized.} 
\end{cases}
\end{equation*}

As shown in Figure \ref{fig:VAE_Arch}, in the encoder, dropout is applied with probability $E_p$ on the output of the hidden layer(s) of the multi-layer feed-forward neural network (FFNN). This output is then fed to two separate  layers to get the approximate posterior $q_W(z|x)$.
In the decoder, dropout is applied with probability $D_p$ on the document-topic distribution vector ($\vec{\theta}_{K \times 1}$), just before the reconstruction process.

\subsection{Task Description}
The goal is to measure the effect that dropout has on the \textit{performance} of VAE-NTMs by varying the dropout rates from $0.0$ to $0.6$ in steps of $0.1$, in both the encoder and the decoder. We have chosen $0.6$ as the upper bound of the dropout rates for our experiments because it is the highest dropout rate used in any VAE-NTMs that we have considered as a baseline in this work. 
We measure performance using: \textit{topic coherence}, \textit{topic diversity}, and \textit{topic quality}. We use NPMI \cite{lau-etal-2014-machine, roder} to measure topic coherence. Topic diversity \cite{ETM} shows the uniqueness of topics. Topic quality is the product of coherence and diversity \cite{ETM}.
As the automated topic model measures do not always accurately capture the quality of the topics \cite{hoyle2021is}, we also perform a manual evaluation of the topics and study their predictive performance on the document classification task.

\begin{figure*}[!htbp]
\centering
\begin{subfigure}[b]{\textwidth}
   \includegraphics[width=1\linewidth]{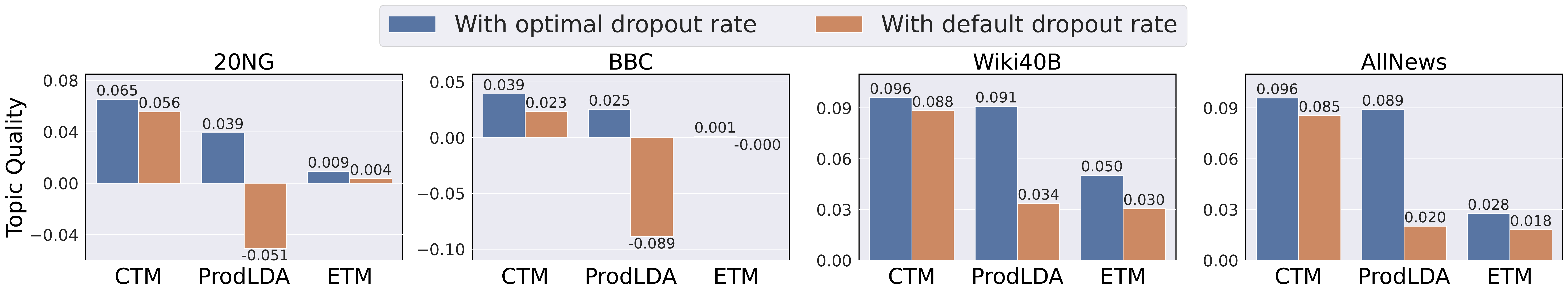}
   \label{fig:TQ} 
\end{subfigure}
\begin{subfigure}[b]{\textwidth}
   \includegraphics[width=1\linewidth]{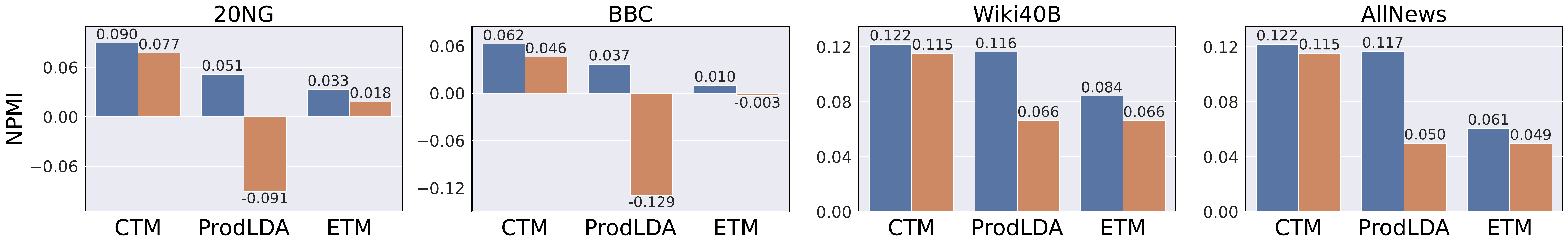}
   \label{fig:NPMI}
\end{subfigure}
\caption{Topic quality and NPMI for different topic models with optimal dropout rate and default dropout rate.
}
\label{fig:compare}
\end{figure*}

\section{Empirical Study}
We perform all experiments in OCTIS \cite{octis}, which is an integrated framework for topic modeling.

\subsection{Datasets}
We have used four publicly available datasets in our experiments. Among them, \textbf{20NG}\footnote{\url{http://qwone.com/~jason/20Newsgroups/}} and \textbf{BBC} \cite{BBCnews} are already available in OCTIS in the pre-processed format while we added \textbf{Wiki40B} \cite{Wiki40B} and \textbf{AllNews} \cite{AllNews} datasets further.
The statistical descriptions of these datasets are mentioned in Table \ref{tab:datasets}. Each corpus is split into train/valid/test sets in the ratio $70\colon15\colon15$. The validation set is used for early stopping. 
\begin{table}[ht]
    \centering
    \begin{adjustbox}{width=.9\linewidth}
      \begin{tabular}{c c c c} \hline \hline
        \multirow{1}{*}{\textbf{Dataset}} &  \textbf{\#Docs}  & \multirow{1}{*}{\textbf{Avg. \#words}} & \multirow{1}{*}{\textbf{|Vocab|}} \\  \hline \hline
        
         \multirow{1}{*}{\textbf{20NG}} & $16309$ & \multirow{1}{*}{$48.02$} & \multirow{1}{*}{$1612$}  \\ 

         \multirow{1}{*}{\textbf{BBC}} & $2225$ & \multirow{1}{*}{$120.12$} & \multirow{1}{*}{$2949$}  \\ 

         \textbf{Wiki40B} & $24774$ & $541.08$ & $2000$  \\

         \textbf{AllNews} & $49754$ & $229.53$ & $2000$  \\ \hline \hline

    \end{tabular}
    \end{adjustbox}
    \caption{Statistics of the used datasets. \label{tab:datasets}}
\end{table}

\subsection{Models}
We use the following three VAE-NTMs: \textbf{CTM} \cite{CombinedTM} which incorporates the contextualized documents embeddings with the neural topic models; \textbf{ProdLDA} \cite{AVITM} which, unlike LDA, relaxes the simplex constraint over the topic-word matrix;  (\textbf{ETM}) \cite{ETM} which incorporates word-embeddings in topic modeling to increase robustness in presence of stopwords.

\begin{table}[ht] 
    \centering
    \begin{adjustbox}{width=\linewidth}
    \begin{tabular}{c c c c c}
    \hline \hline
         \textbf{Model} &\textbf{20NG} & \textbf{BBC} & \textbf{Wiki40B} & \textbf{AllNews}\\ \hline \hline
         \makecell{\textbf{CTM}\\$(0.2,\: 0.2)$} & $(0.0,\: 0.0)$ & $(0.0, \: 0.0)$ & $(0.2,\: 0.1)$ & $(0.0,\: 0.1)$ \\ \hline
                              
         \makecell{\textbf{ProdLDA}\\$(0.6,\: 0.6)$} & $(0.1,\: 0.1)$ & $(0.0,\: 0.0)$ & $(0.1,\: 0.1)$ & $(0.1,\: 0.1)$ \\ \hline
                              
         \makecell{\textbf{ETM}\\$(0.5,\: 0.0)$} & $(0.0,\: 0.0)$ & $(0.1,\: 0.0)$ & $(0.0,\: 0.0)$ & $(0.1,\: 0.0)$ \\ \hline \hline
    \end{tabular}
    \end{adjustbox}
    \caption{For each of the datasets, the optimal dropout rates of all the models considering the highest topic quality are mentioned in the $({E_p}, {D_p})$ format in the second through last columns. The default dropout rate is also specified for each model in the first column.
    \label{tab:OptimizedDrop}
    }
\end{table}

For each of the four datasets, we compute the dropout rate that optimizes  the topic quality of each model on that dataset. 
We train all three topic models for topic-count $K \in \{20, 50, 100\}$ with 30 epochs while keeping all hyperparameter values, except dropout, the same as in their original implementations. To ensure robustness, we average scores over 10 independent runs of each model. For comparison, we use the default  dropout rates for each model as mentioned in the original papers that proposed the corresponding model. In Table \ref{tab:OptimizedDrop}, we show the default and the optimal dropout rates. 

\begin{figure*}[ht]
    \centering
    \includegraphics[width=\linewidth]{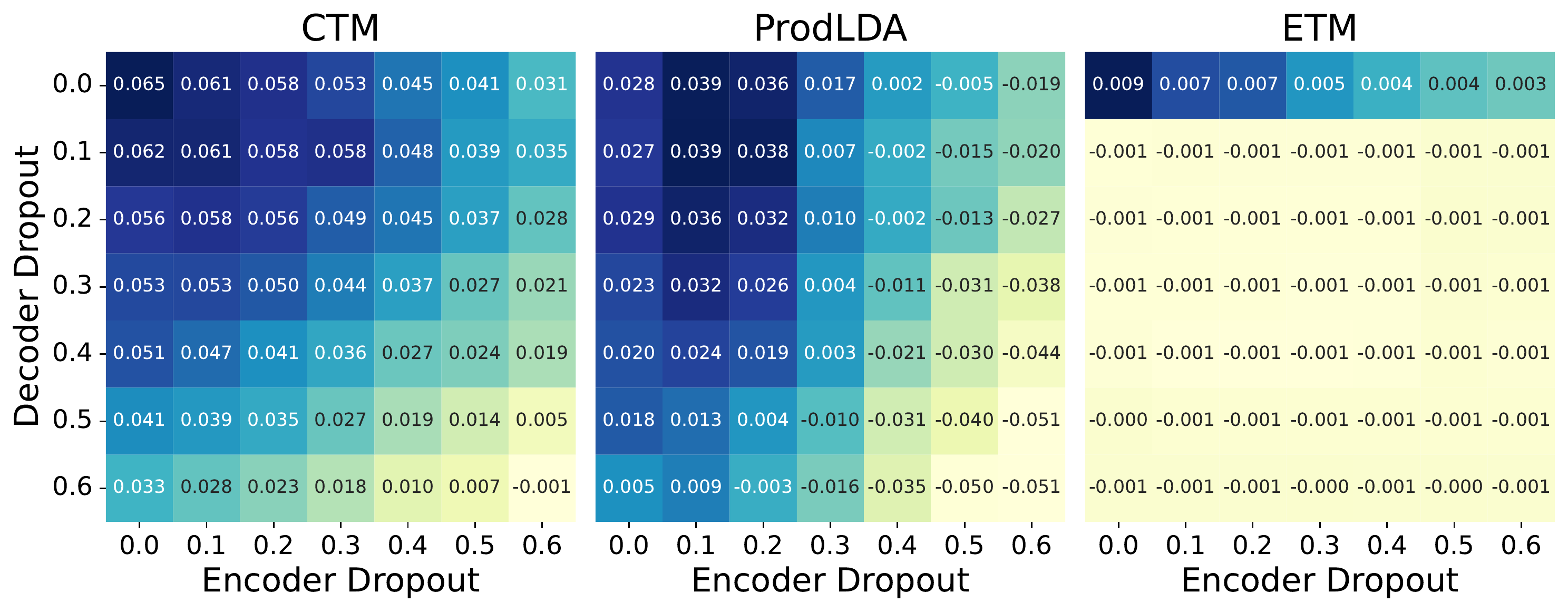}
    \caption{Topic qualities on \textbf{20NG} for $(E_p, D_p) \in [0.0,\;0.6] \times [0.0,\;0.6]$ with a increment of $0.1$.}
    \label{fig:TopicQuality}
\end{figure*}

\subsection{Results and Analysis}

\subsubsection{Quantitative Evaluation of Topic Quality}
In Figure \ref{fig:compare}, we compare, for each dataset and each model, the topic quality and the NPMI respectively between the dropout-optimized model that gives the highest topic quality and the model with default dropout rates as mentioned in Table \ref{tab:OptimizedDrop}. 

On \textbf{20NG}, the topic quality score for (CTM, ProdLDA, ETM) is improved from $(0.056, -0.051, 0.004)$ to $(0.065, 0.039, 0.009)$ by optimizing the dropout rate. For CTM, the increase in performance is around $16.07\%$ whereas for the other two models it is over $100\%$. This is because the original implementation of CTM already uses a relatively low dropout rate, i.e., $0.2$, for both the encoder and the decoder. The other two models show a significant increase in performance due to their large dropout in the baseline models. 

Figure \ref{fig:TopicQuality} shows that the topic quality on the \textbf{20NG} dataset for the VAE-NTMs generally produces better results on keeping the dropout rate for both the encoder and the decoder either to be zero or close to it, especially values like $\{0.0, 0.1\}$. Similar results have been found for the other datasets. Based on these observations, the topic quality is found to reduce with an increase in dropout rates in the encoder and decoder.

\subsubsection{Qualitative Evaluation of Topic Quality}
To qualitatively evaluate the models, we trained all of them for a topic count of $100$ on the \textbf{20NG} dataset. We then aligned the topics for each pair of \textit{(optimal-dropout model, default-dropout model)} for all three different models. We followed a two-step strategy for topic alignment. For a given pair of models, namely, one with optimal dropout and another with default dropout, with topics lists $P$ and $Q$, respectively,   we first construct a similarity matrix of the topic lists using Rank-biased Overlap \cite{Webber2010RBO} (RBO) which computes the similarity between two ordered lists by taking into consideration the rank of the individual elements. For example, for $100$ topics, we get a matrix, $\mathbf{A}=(a_{ij})_{1 \leq i,j \leq 100}$ such that, $a_{i,j} = \operatorname{RBO}\big(P[i], Q[j]\big)$. The RBO score lies in $[0, 1]$, where $0$ represents no overlap and $1$ implies exact overlap. In the final step, we iteratively select the pair of topics for which the similarity score is maximum and simultaneously exclude these two topics from further consideration, i.e. if $\big(P[i_1], Q[j_1]\big)$ and $\big(P[i_2], Q[j_2]\big)$ are two selected pairs then $\left(i_1 \neq i_2 \land j_1 \neq j_2\right)$.

In Table \ref{tab:topics_20NG} we show the top words from \textit{aligned} topics of all the models. `$\ast$' marked models have dropout optimized to give the highest topic quality while others use the default dropout rates as mentioned in Table \ref{tab:OptimizedDrop}. We see that dropout-optimized models output more  interpretable topics.
\begin{table*}[ht]
\centering
\begin{adjustbox}{width=.96\linewidth}
  \begin{tabular}{  c  l  } \hline \hline
  \textbf{Model} & \multicolumn{1}{c}{\textbf{Topics}} \\ \hline \hline
     \multirow{3}{*}{\makecell{\textbf{CTM*}\\${(0.0, 0.0)}$}} 
         & \textbf{monitor}, \textbf{card}, \textbf{video}, \textbf{port}, \textbf{vga}, \textbf{apple}, \textbf{connector}, \textbf{serial}, \textbf{slot}, \textbf{output}\\ 
         & \textbf{firearm}, \textbf{weapon}, \textbf{dangerous}, \textbf{military}, \textbf{license}, \textit{file}, \textit{state}, \textbf{gun}, \textbf{police}, \textit{issue} \\ 
         & \textbf{christian}, \textbf{truth}, \textbf{scripture}, \textit{exist}, \textbf{belief}, \textbf{accept}, \textit{understand}, \textit{word}, \textbf{human}, \textbf{doctrine}\\ \hline 
         
     \multirow{3}{*}{\makecell{\textbf{CTM}\\${(0.2, 0.2)}$}}
         & \textbf{card}, \textbf{monitor}, \textbf{video}, \textit{offer}, \textit{sale}, \textit{upgrade}, \textbf{mouse}, \textbf{vga}, \textbf{port}, \textbf{parallel}\\ 
         & \textbf{firearm}, \textbf{dangerous}, \textbf{license}, \textbf{weapon}, \textit{section}, \textit{file}, \textit{division}, \textit{device}, \textit{manufacture}, \textit{carry} \\
         & \textit{interpretation}, \textbf{truth}, \textbf{scripture}, \textbf{christian}, \textbf{agree}, \textbf{moral}, \textit{understand}, \textbf{human}, \textbf{faith}, \textit{claim} \\ \hline \hline

     \multirow{3}{*}{\makecell{\textbf{ProdLDA*}\\${(0.1, 0.1)}$}}
         & \textbf{window}, \textbf{driver}, \textit{mode}, \textbf{run}, \textbf{mouse}, \textbf{session}, \textbf{server}, \textbf{program}, \textbf{manager}, \textbf{install}\\
         & \textbf{car}, \textbf{engine}, \textbf{buy}, \textbf{company}, \textbf{vehicle}, \textit{make}, \textbf{brake}, \textbf{tire}, \textbf{dealer}, \textbf{road}\\ 
         & \textbf{signal}, \textbf{voltage}, \textbf{output}, \textbf{circuit}, \textbf{noise}, \textbf{power}, \textbf{switch}, \textbf{wire}, \textbf{connector}, \textit{degree}\\  \hline 
         
     \multirow{3}{*}{\makecell{\textbf{ProdLDA}\\${(0.6, 0.6)}$}}
         & \textit{line}, \textbf{window}, \textit{gun}, \textbf{read}, \textbf{space}, \textbf{run}, \textit{statement}, \textbf{datum}, \textbf{drive}, \textit{make}\\
         & \textit{make}, \textbf{battery}, \textbf{engine}, \textit{homosexual}, \textit{assault}, \textit{reason}, \textit{place}, \textit{single}, \textit{large}, \textit{attempt}\\
         & \textbf{voltage}, \textit{damn}, \textbf{signal}, \textit{usual}, \textit{label}, \textit{hour}, \textit{bio}, \textit{leg}, \textit{bullet}, \textit{hundred}\\ \hline \hline

     \multirow{3}{*}{\makecell{\textbf{ETM*}\\${(0.0, 0.0)}$}}
         & \textbf{version}, \textbf{software}, \textbf{program}, \textbf{file}, \textit{include}, \textbf{image}, \textbf{application}, \textit{set}, \textbf{server}, \textit{support} \\ 
         & \textbf{armenian}, \textbf{turkish}, \textbf{village}, \textbf{people}, \textbf{israeli}, \textbf{population}, \textbf{muslim}, \textbf{kill}, \textbf{russian}, \textbf{genocide} \\
         & \textbf{system}, \textbf{run}, \textbf{work}, \textbf{window}, \textit{problem}, \textit{include}, \textit{set}, \textit{good}, \textit{support}, \textbf{information}\\ \hline

     \multirow{3}{*}{\makecell{\textbf{ETM}\\${(0.5, 0.0)}$}}
         & \textbf{file}, \textbf{application}, \textit{set}, \textbf{program}, \textit{support}, \textbf{image}, \textbf{display}, \textit{list}, \textbf{version}, \textbf{bit}\\ 
         & \textbf{armenian}, \textbf{turkish}, \textbf{village}, \textbf{israeli}, \textbf{population}, \textbf{muslim}, \textbf{genocide}, \textit{son}, \textit{land}, \textbf{jewish}\\ 
         & \textbf{work}, \textit{call}, \textbf{system}, \textbf{window}, \textit{problem}, \textbf{bit}, \textit{set}, \textbf{run}, \textit{support}, \textit{good} \\ \hline \hline

\end{tabular}
\end{adjustbox}
\caption{Some selected topics among 100 topics from \textbf{20NG}. `*' indicates models with optimal dropout. The dropout rate is mentioned in the $(E_p, D_p)$ format. The more related words in a topic are highlighted in bold while less related ones are italicized.
\label{tab:topics_20NG}}
\end{table*}

\subsubsection{Effect of Dataset Length}
Among the input datasets on which we have experimented, the \textbf{20NG} dataset contains relatively short texts, while the others contain longer texts. \cite{LDAdropout} find that their dropout methods are not effective on long texts. But here we see that the performance of all VAE-NTMs decreases uniformly with the increase in the dropout rate, irrespective of the length of the dataset.

\subsubsection{Document Classification}
\begin{figure}[ht]
    \centering
    \includegraphics[width=\linewidth]{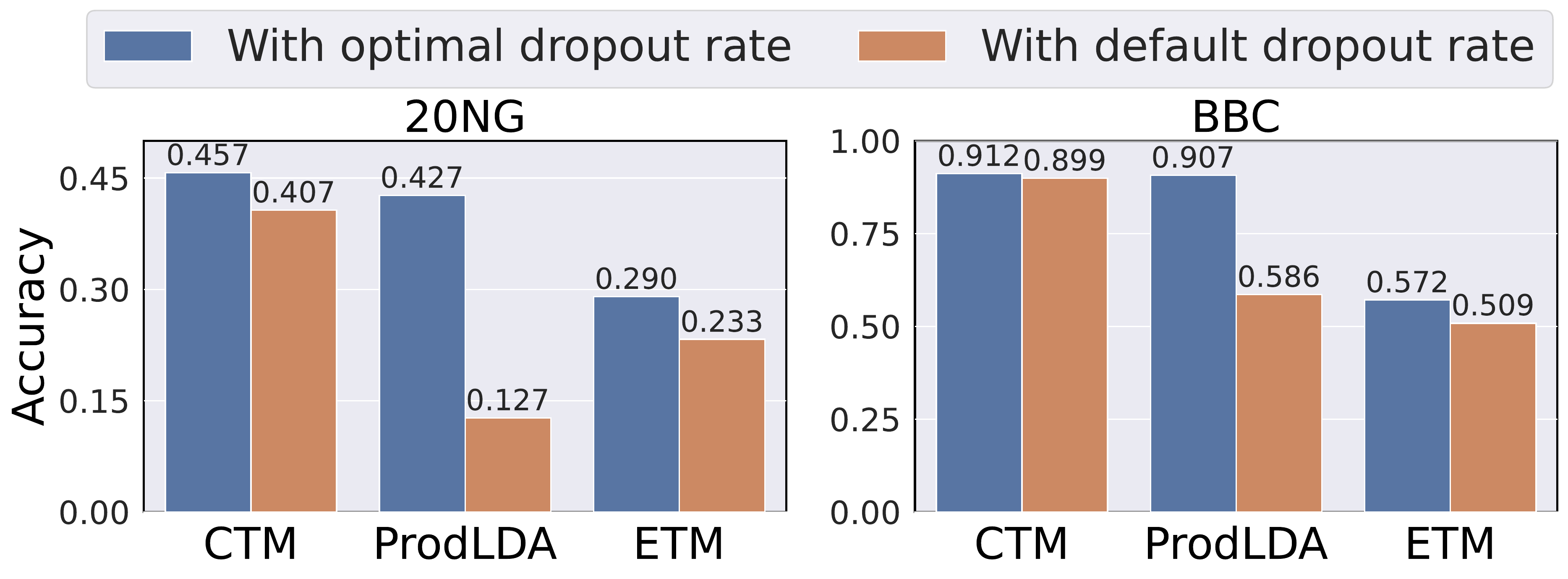}
    \caption{Accuracy for different topic models with optimal dropout and default dropout from Table \ref{tab:OptimizedDrop}.}
    \label{fig:Acc_def}
\end{figure}

We test the predictive performance of the topics produced by the models on a document classification task. We train the models on \textbf{20NG} and \textbf{BBC} corpora for $K$ topics using the training subset. We represent each document as a $K$-dimensional document-topic vector and train an SVM, which is then tested on the test subset. We average the accuracy scores over $K \in \{20,50,100\}$.
Figure \ref{fig:Acc_def} shows that accuracy increases when we use the optimized dropout rates. 
\section{Theoretical Understanding of Results}
Our experiments show that by tuning the dropout carefully, we can achieve a significant improvement in the performance of VAE-NTMs. Therefore, we argue that the dropout rate should be treated as an important hyperparameter and 
carefully selected based on the choice of the model as well as the dataset, especially in the case of VAE-NTMs. More precisely, in most cases, low dropout rates in the encoder and the decoder lead to higher performance than that achieved for higher dropout rates.

Standard dropout and other types of dropout have been extensively used in supervised learning techniques \cite{SrivastavaDropout, TowardsDropCNN, DropCNN, ImprovedDropCNN, EEDropCNN}. The main prerogative of using dropout in the supervised scenario is to introduce noise while training so that the model can recognize the outliers in the testing phase. The drop in performance with high dropout that we see in our experiments is perhaps due to the fact that we are trying to learn a  generative model of the data. Dropout makes the model robust against perturbations in the input data and thereby also prevents it from learning the characteristics of the input distribution accurately. This is probably why we see a drop in topic coherence and quality. In the case of document classification, if the topic model is trained with a high dropout, the document-topic vectors are of poor quality and the classifier gets trained on these vectors; this results in poor accuracy on the test documents. This setting is different from the usual supervised learning of neural classifiers where dropout is introduced directly in the classifier to prevent overfitting. We intend to analyze these aspects in more depth in the future.

\section{Conclusion}
We present a detailed study of the effect of the dropout rate on VAE-NTMs. We find that the model performance generally reduces with the increase in dropout rate in the encoder as well as the decoder.

\section*{Limitations}
The following limitations are known and should be considered when applying the results of this work or relying on them in future studies:
(1) Other variants of dropout can be applied to the VAE-NTMs.
(2) Analysis of the dropout effect may be done for other VAE-NTMs as well.
(3) Other downstream tasks may be formulated for further analysis.

\section*{Acknowledgments}
This work is partially supported by the SERB-DST Project CRG/2021/000803 sponsored by the Department of Science and Technology, Government of India at Indian Association for the Cultivation of Science, Kolkata.

\appendix

\section{Appendix}
\label{sec:appendix}
\subsection{Datasets}
\label{sec:ap_Datasets}
We run our experiments on the following datasets: 
\begin{itemize}
    \item \textbf{20NewsGroups (20NG)}
     is a dataset of $18,846$ documents from $20$ different newsgroups posts. The \textbf{20NG} dataset is present in OCTIS, so it is already in pre-processed form. All the documents of this dataset have their corresponding category type as the document labels. The details about these categories are mentioned in Table \ref{tab:20NG_labels}.
    
    \item \textbf{BBC News (BBC)} \cite{BBCnews} is a dataset of news articles from BBC. It is also accessible from OCTIS in pre-processed form. The documents of this dataset are categorized into 5 different categories which are \textit{tech}, \textit{business}, \textit{entertainment}, \textit{sports}, and \textit{politics}. The details of these categories are mentioned in Table \ref{tab:BBC_labels}.
    
    \item \textbf{Wiki40B}
    \cite{Wiki40B} is a Wikipedia text dataset in 40\texttt{+} languages, available in TensorFlow dataset format. In our experiment, we take a sample of $24,774$ English documents from this dataset.
    
    \item \textbf{All the News (AllNews)}
    \cite{AllNews} dataset consists of $50,001$ news articles from $15$ news publishers.
\end{itemize}
   
\begin{table}[!htbp]
\centering
\begin{adjustbox}{width=\linewidth}
  \begin{tabular}{ c  c  c  c } \hline \hline
    \textbf{\#No.} & \textbf{Label} & \textbf{\#Docs} & \textbf{\%Docs} \\ \hline \hline
    1. & misc.forsale & 861 & 5.28 \\
    2. & comp.windows.x & 883 & 5.41 \\
    3. & soc.religion.christian & 920 & 5.64 \\
    4. & talk.religion.misc & 521 & 3.19 \\
    5. & rec.autos & 822 & 5.04 \\
    6. & sci.med & 866 & 5.31 \\
    7. & talk.politics.misc & 689 & 4.22 \\
    8. & talk.politics.mideast & 828 & 5.08 \\
    9. & sci.electronics & 867 & 5.32 \\
    10. & rec.sport.hockey & 843 & 5.17 \\
    11. & rec.sport.baseball & 787 & 4.83 \\
    12. & talk.politics.guns & 808 & 4.95 \\
    13. & sci.crypt & 883 & 5.41 \\
    14. & comp.sys.mac.hardware & 838 & 5.14 \\
    15. & comp.sys.ibm.pc.hardware & 891 & 5.46 \\
    16. & comp.graphics & 836 & 5.13 \\
    17. & comp.os.ms-windows.misc & 828 & 5.08 \\
    18. & alt.atheism & 689 & 4.22 \\
    19. & sci.space & 856 & 5.25 \\
    20. & rec.motorcycles & 793 & 4.86 \\ \hline \hline
\end{tabular}
\end{adjustbox}
\caption{\textbf{20NG} labels with corresponding document counts and percentage of documents.\label{tab:20NG_labels}}
\end{table}

\begin{table}[!htbp]
\centering
\begin{adjustbox}{width=.85\linewidth}
  \begin{tabular}{c c c c } \hline \hline
    \textbf{\#No.} & \textbf{Label} & \textbf{\#Docs} & \textbf{\%Docs} \\ \hline \hline
    1. & tech & 401 & 18.02 \\
    2. & business & 510 & 22.92 \\
    3. & entertainment & 386 & 17.35 \\
    4. & sport & 511 & 22.97 \\
    5. & politics & 417 & 18.74 \\ \hline \hline
\end{tabular}
\end{adjustbox}
\caption{\textbf{BBC} labels with corresponding document counts and percentage of documents.\label{tab:BBC_labels}}
\end{table}

\subsection{Pre-processing Steps}
\label{sec:ap_preprocess}
Using OCTIS, we convert each document to lowercase, remove the punctuations, lemmatize it, filter the vocabulary with the most frequent 2000 terms, filter words with less than 3 characters, and filter documents with less than 3 words.

\subsection{Topic Evaluation Metrics}
\label{sec:ap_TopicEval}
\begin{enumerate}
    \item \textbf{Coherence metric}: This measures how much the top words of the topics are relevant. Topic coherence $(\operatorname{TC})$ for $K$ topics each of which contains $n$ top words can be calculated as:
    \begin{equation*}
        \operatorname{TC} = \frac{1}{K} \sum_{k=1}^K \frac{1}{\Mycomb[n]{2}}\sum_{i=1}^n \sum_{j=i+1}^n f\left(w_i^{(k)}, w_j^{(k)}\right)
    \end{equation*}
    Here, $f(\cdot, \cdot)$ is the Normalized Pointwise Mutual Information or NPMI \cite{lau-etal-2014-machine} of the words $w_i ^{(k)}$ and $w_j ^{(k)}$ appearing in topic $k$:
    \begin{equation*}
        f(w_i^{(k)}, w_j^{(k)}) = \frac{\log{\frac{p(w_i^{(k)}, w_j^{(k)}) + \epsilon}{p(w_i^{(k)})p(w_j^{(k)})}}}{- \log{\left(p(w_i^{(k)}, w_j^{(k)})+\epsilon\right)}}
    \end{equation*}
    where, $p(w_i^{(k)}, w_j^{(k)})$ is the probability of the co-occurrence of the words $w_i^{(k)}$ and $w_j^{(k)}$ in a boolean sliding window in topic $k$ and $p(w_i^{(k)})$ and $p(w_j^{(k)})$ represents the probability of the occurrence of the individual words in topic $k$. $\epsilon$ is a small positive constant that is used to avoid zero in the $\log(\cdot)$ function. 

    \item \textbf{Diversity metric}: This measures how much the generated topics are different from each other. To measure the diversity score we have used the metric Topic Diversity (TD) \cite{ETM} which is defined as the proportion of the number of unique words appearing across all topics.
    It ranges between $[0, 1]$ where a value close to $0$ implies repetitive topics and a value near $1$ represents more diversification in the topics.

    \item \textbf{Topic quality}: This is an overall metric that is defined as the product of the two metrics NPMI and TD.
\end{enumerate}

In our experiments, we take the top 10 words for each topic (i.e., $n=10$) to compute NPMI and TD scores.


\subsection{Detailed Results}
\label{sec:ap_Detailed_Results}
The detailed results of our experiments are given in Tables \ref{tab:tq_results}, \ref{tab:npmi_results}, and \ref{tab:TD_results}. An asterisk ($\ast$) against a model in the above tables indicates that it is trained with the optimal dropout rate, and the absence of an asterisk indicates that the default dropout rate is used. The default dropout rate for \textbf{CTM} is taken from \cite{CombinedTM}, for \textbf{ProdLDA} from \cite{AVITM}, and for \textbf{ETM} from \cite{ETM}.
\begin{table}[ht]
    \centering
    \begin{adjustbox}{width=\linewidth}
      \begin{tabular}{c c c c c} \hline \hline
        \multirow{2}{*}{\textbf{Model}} &  \multicolumn{4}{c}{\textbf{Topic quality for each dataset}} \\  \cline{2-5}
        & \textbf{20NG} & \textbf{BBC} & \textbf{Wiki40B} & \textbf{AllNews} \\ \hline \hline
        \textbf{CTM*} & \makecell{0.0652} & \makecell{0.0392} & \makecell{0.0961} & \makecell{0.0958}
        \\ 
        \makecell{\textbf{CTM}} & 0.0556 & 0.0234 & 0.0884 & 0.0854
        
           \\ \hline 
           
        \textbf{ProdLDA*} & \makecell{0.0392} & \makecell{0.0254} & \makecell{0.0910}  & \makecell{0.0891}
        \\ 
        \makecell{\textbf{ProdLDA}} & -0.0507 & -0.0887 & 0.0336 & 0.0202
        
         \\ \hline 

         \textbf{ETM*} & \makecell{0.0092} & \makecell{0.0011} & \makecell{0.0502} & \makecell{0.0276}
        \\ 
        \makecell{\textbf{ETM}} & 0.0036 & -0.0003 & 0.0304 & 0.0181
        
         \\ \hline \hline
         
    \end{tabular}
    \end{adjustbox}
    \caption{Topic Quality values for different VAE-NTMs with optimal dropout rate and default dropout rate (see Table \ref{tab:OptimizedDrop}). `*' indicates models with optimal dropout.
    \label{tab:tq_results}}
\end{table}

\begin{table}[ht]
    \centering
    \begin{adjustbox}{width=\linewidth}
      \begin{tabular}{c c c c c} \hline \hline
        \multirow{2}{*}{\textbf{Model}} &  \multicolumn{4}{c}{\textbf{NPMI for each dataset}} \\  \cline{2-5}
        & \textbf{20NG} & \textbf{BBC} & \textbf{Wiki40B} & \textbf{AllNews} \\ \hline \hline
        \textbf{CTM*} & \makecell{0.0896} & \makecell{0.0623} & \makecell{0.1219} & \makecell{0.1218}
        \\
        \makecell{\textbf{CTM}} & 0.0774 & 0.0458 & 0.1152 & 0.1153
        
           \\ \hline 
           
        \textbf{ProdLDA*} & \makecell{0.0513} & \makecell{0.0367} & \makecell{0.1162}  & \makecell{0.1166}
        \\ 
        \makecell{\textbf{ProdLDA}} & -0.0907 & -0.1293 & 0.0662 & 0.0498
        
         \\ \hline

         \textbf{ETM*} & \makecell{0.0331} & \makecell{0.0100} & \makecell{0.0841} & \makecell{0.0605}
        \\
        \makecell{\textbf{ETM}} & 0.0183 & -0.0033 & 0.0662 & 0.0494
        
         \\ \hline \hline
         
    \end{tabular}
    \end{adjustbox}
    \caption{NPMI values for different VAE-NTMs with optimal dropout rate and default dropout rate (see Table \ref{tab:OptimizedDrop}). `*' indicates models with optimal dropout.
    \label{tab:npmi_results}}
\end{table}

\begin{table}[ht]
    \centering
    \begin{adjustbox}{width=\linewidth}
      \begin{tabular}{c c c c c} \hline \hline
        \multirow{2}{*}{\textbf{Model}} &  \multicolumn{4}{c}{\textbf{TD for each dataset}} \\  \cline{2-5}
        & \textbf{20NG} & \textbf{BBC} & \textbf{Wiki40B} & \textbf{AllNews} \\ \hline \hline
        \textbf{CTM*} & \makecell{0.7283} & \makecell{0.6295} & \makecell{0.7883} & \makecell{0.7871}
        \\ 
        \makecell{\textbf{CTM}} & 0.7175 & 0.51 & 0.7671 & 0.7409
        
           \\ \hline 
           
        \textbf{ProdLDA*} & \makecell{0.7644} & \makecell{0.6902} & \makecell{0.7829}  & \makecell{0.7640}
        \\ 
        \makecell{\textbf{ProdLDA}} & 0.5594 & 0.6861 & 0.5071 & 0.4061
        
         \\ \hline

         \textbf{ETM*} & \makecell{0.2776} & \makecell{0.1108} & \makecell{0.5973} & \makecell{0.4561}
        \\ 
        \makecell{\textbf{ETM}} & 0.1949 & 0.0902 & 0.4599 & 0.3659
        
         \\ \hline \hline
         
    \end{tabular}
    \end{adjustbox}
    \caption{Topic Diversity values for different VAE-NTMs with optimal dropout rate and default dropout rate (see Table \ref{tab:OptimizedDrop}). `*' indicates models with optimal dropout.
    \label{tab:TD_results}}
\end{table}

\begin{figure*}[ht]
\centering
\begin{subfigure}[b]{\textwidth}  
    \centering 
    \includegraphics[width=\textwidth]{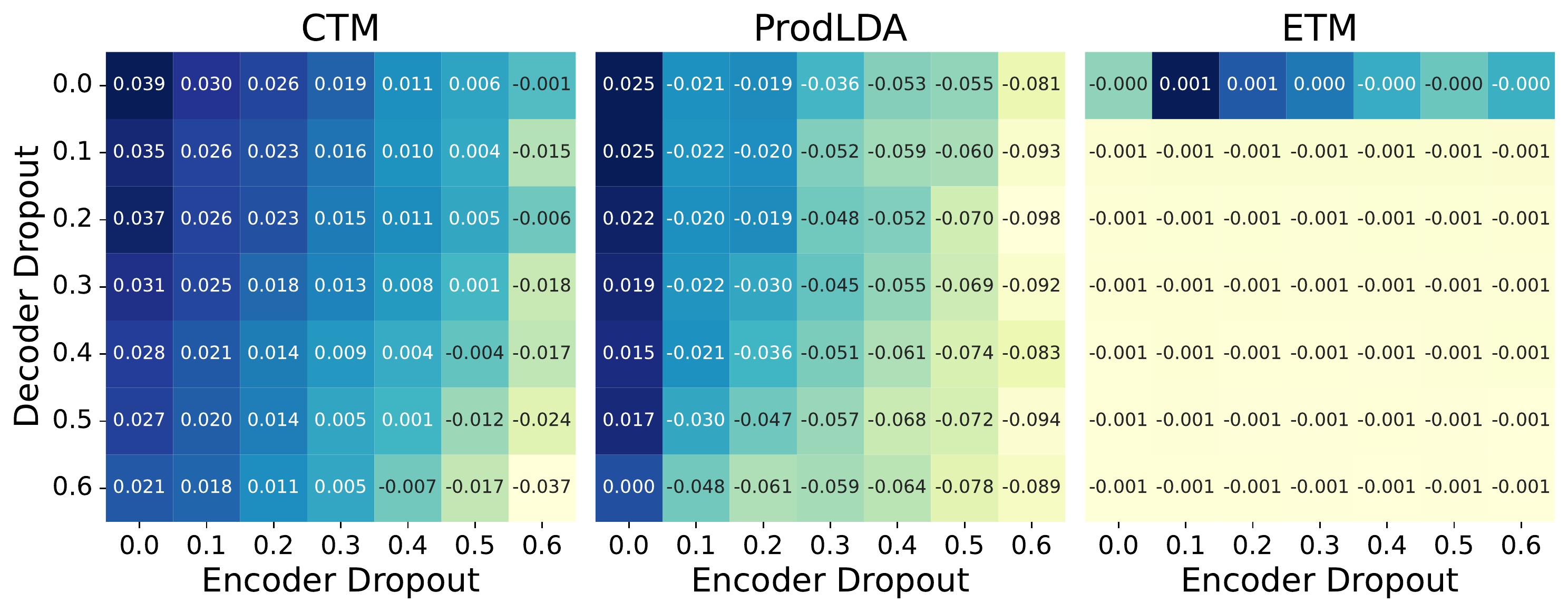}
    \caption{{\small \texttt{BBC}}}    
    \label{fig:BBC_Heatmap_TopicQulaity}
\end{subfigure}
\vskip\baselineskip
\begin{subfigure}[b]{\textwidth}   
    \centering 
    \includegraphics[width=\textwidth]{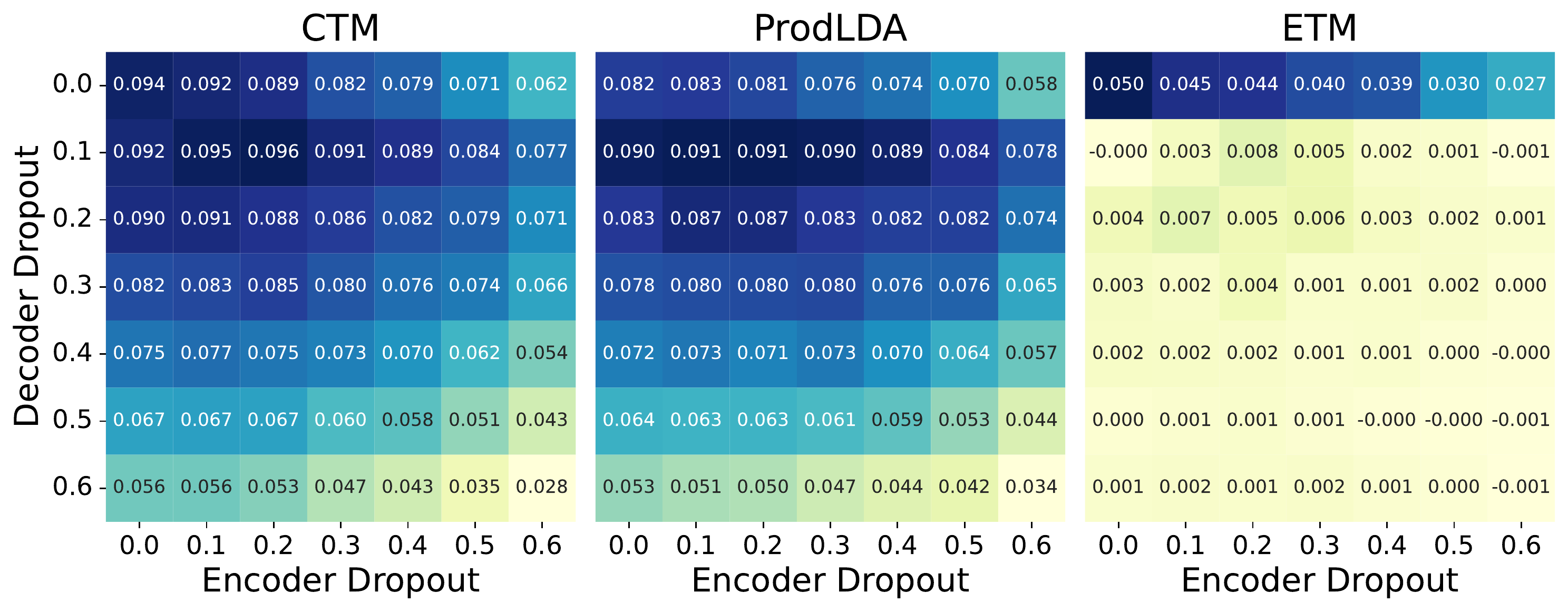}
    \caption{{\small \texttt{Wiki40B}}}    
    \label{fig:Wiki40B_Heatmap_TopicQulaity}
\end{subfigure}
\vskip\baselineskip
\begin{subfigure}[b]{\textwidth}   
    \centering 
    \includegraphics[width=\textwidth]{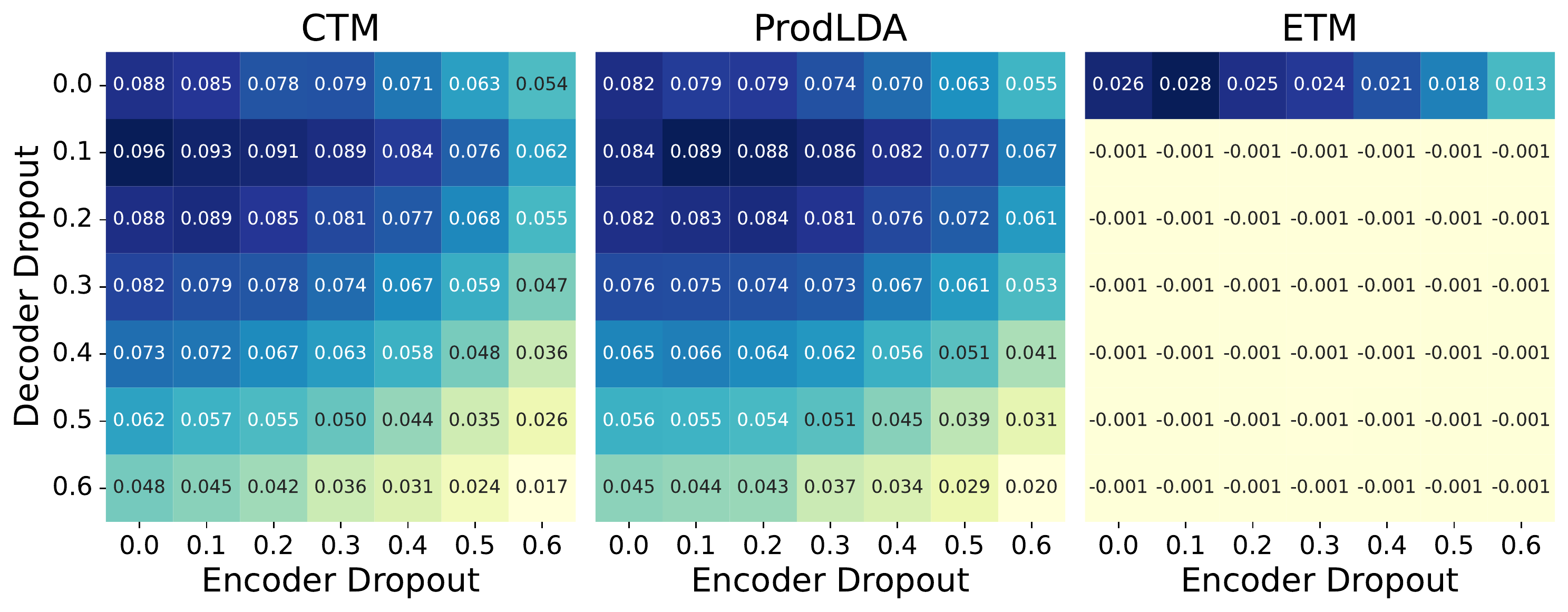}
    \caption{{\small \texttt{AllNews}}}    
    \label{fig:AllNews_Heatmap_TopicQulaity}
\end{subfigure}
\caption{\small Change in \textbf{topic quality} for $(E_p, D_p) \in [0.0,\;0.6] \times [0.0,\;0.6]$ with a increment of $0.1$.} 
\label{fig:TopicQulaity}
\end{figure*}

\begin{figure*}[ht]
\centering
\begin{subfigure}[b]{\textwidth}
    \centering
    \includegraphics[width=\textwidth]{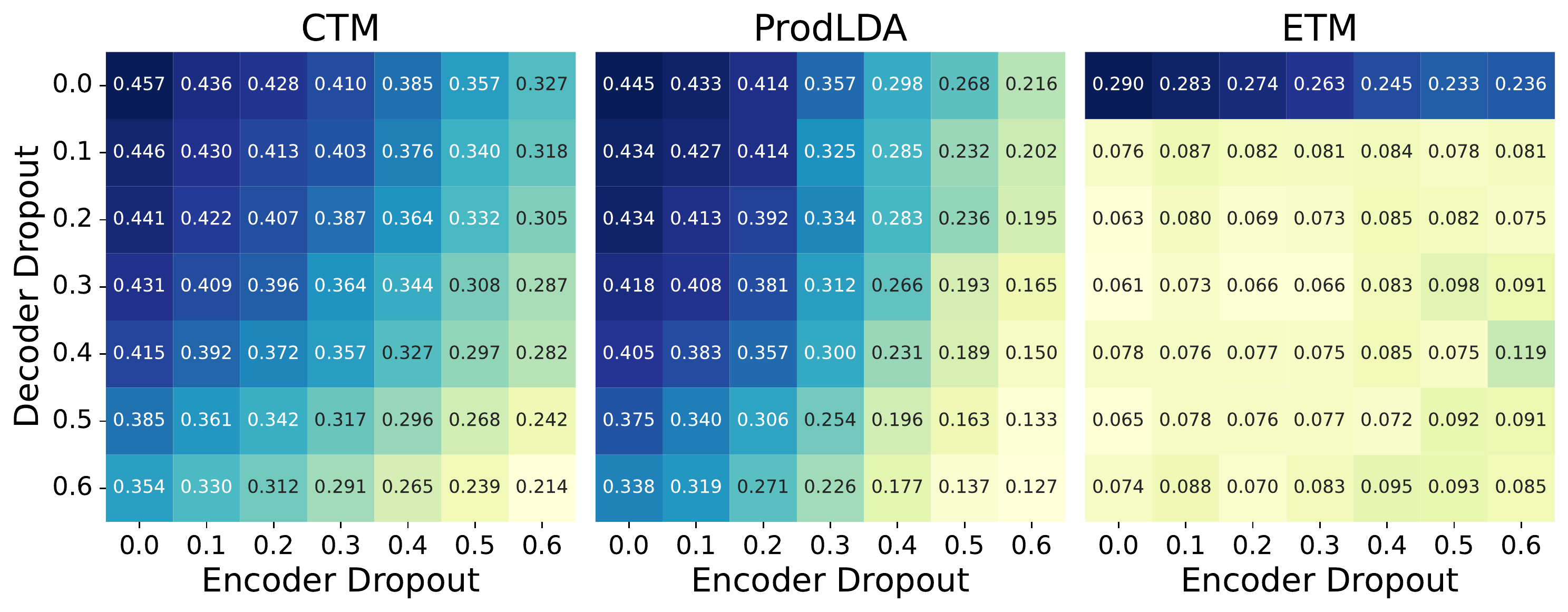}
    \caption{{\small \texttt{20NG}}}    
    \label{fig:20NG_Heatmap_Accuracy}
\end{subfigure}
\vskip\baselineskip
\begin{subfigure}[b]{\textwidth}  
    \centering 
    \includegraphics[width=\textwidth]{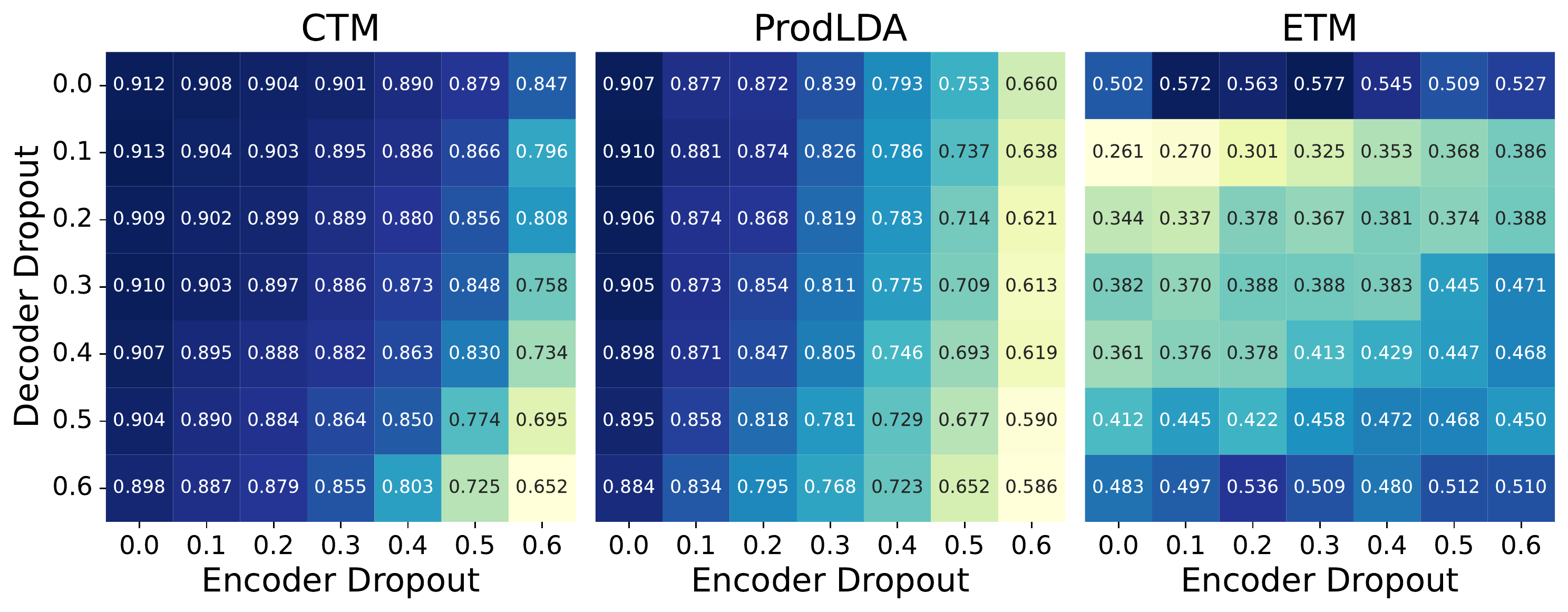}
    \caption{{\small \texttt{BBC}}}    
    \label{fig:BBC_Heatmap_Accuracy}
\end{subfigure}
\caption{\small Accuracy scores for $(E_p, D_p) \in [0.0,\;0.6] \times [0.0,\;0.6]$ (step $=0.1$) in document classification task.} 
\label{fig:Accuracy}
\end{figure*}


\begin{thebibliography}{27}
\expandafter\ifx\csname natexlab\endcsname\relax\def\natexlab#1{#1}\fi

\bibitem[{Adhya and Sanyal(2022)}]{adhya-sanyal-2022-indian}
Suman Adhya and Debarshi~Kumar Sanyal. 2022.
\newblock \href {https://aclanthology.org/2022.politicalnlp-1.10} {What does
  the {I}ndian {P}arliament discuss? an exploratory analysis of the question
  hour in the {L}ok {S}abha}.
\newblock In \emph{Proceedings of the LREC 2022 workshop on Natural Language
  Processing for Political Sciences}, pages 72--78, Marseille, France. European
  Language Resources Association.

\bibitem[{Baldi and Sadowski(2013)}]{baldi_understanding_dropout}
Pierre Baldi and Peter~J Sadowski. 2013.
\newblock \href
  {https://proceedings.neurips.cc/paper/2013/file/71f6278d140af599e06ad9bf1ba03cb0-Paper.pdf}
  {Understanding dropout}.
\newblock In \emph{Advances in Neural Information Processing Systems},
  volume~26. Curran Associates, Inc.

\bibitem[{Bianchi et~al.(2021)Bianchi, Terragni, and Hovy}]{CombinedTM}
Federico Bianchi, Silvia Terragni, and Dirk Hovy. 2021.
\newblock \href {https://doi.org/10.18653/v1/2021.acl-short.96} {Pre-training
  is a hot topic: Contextualized document embeddings improve topic coherence}.
\newblock In \emph{Proceedings of the 59th Annual Meeting of the Association
  for Computational Linguistics and the 11th International Joint Conference on
  Natural Language Processing (Volume 2: Short Papers)}, pages 759--766.
  Association for Computational Linguistics.

\bibitem[{Blei et~al.(2003)Blei, Ng, and Jordan}]{LDA}
David~M Blei, Andrew~Y Ng, and Michael~I Jordan. 2003.
\newblock \href {https://www.jmlr.org/papers/volume3/blei03a/blei03a.pdf}
  {Latent {Dirichlet} allocation}.
\newblock \emph{Journal of Machine Learning Research}, 3(Jan):993--1022.

\bibitem[{Cai et~al.(2019)Cai, Gao, Zhang, Wang, Chen, and Ooi}]{EEDropCNN}
Shaofeng Cai, Jinyang Gao, Meihui Zhang, Wei Wang, Gang Chen, and Beng~Chin
  Ooi. 2019.
\newblock \href {http://arxiv.org/abs/1904.03392} {Effective and efficient
  dropout for deep convolutional neural networks}.
\newblock \emph{CoRR}, abs/1904.03392.

\bibitem[{Devries and Taylor(2017)}]{ImprovedDropCNN}
Terrance Devries and Graham~W. Taylor. 2017.
\newblock \href {http://arxiv.org/abs/1708.04552} {Improved regularization of
  convolutional neural networks with cutout}.
\newblock \emph{CoRR}, abs/1708.04552.

\bibitem[{Dieng et~al.(2020)Dieng, Ruiz, and Blei}]{ETM}
Adji~B. Dieng, Francisco J.~R. Ruiz, and David~M. Blei. 2020.
\newblock \href {https://doi.org/10.1162/tacl_a_00325} {Topic modeling in
  embedding spaces}.
\newblock \emph{Transactions of the Association for Computational Linguistics},
  8:439--453.

\bibitem[{Greene and Cunningham(2006)}]{BBCnews}
Derek Greene and P\'{a}draig Cunningham. 2006.
\newblock \href {https://doi.org/10.1145/1143844.1143892} {Practical solutions
  to the problem of diagonal dominance in kernel document clustering}.
\newblock In \emph{Proceedings of the 23rd International Conference on Machine
  Learning}, ICML '06, page 377–384, New York, NY, USA. Association for
  Computing Machinery.

\bibitem[{Guo et~al.(2020)Guo, Dai, Vrande{\v{c}}i{\'c}, and Al-Rfou}]{Wiki40B}
Mandy Guo, Zihang Dai, Denny Vrande{\v{c}}i{\'c}, and Rami Al-Rfou. 2020.
\newblock \href {https://aclanthology.org/2020.lrec-1.297} {{W}iki-40{B}:
  Multilingual language model dataset}.
\newblock In \emph{Proceedings of the 12th Language Resources and Evaluation
  Conference}, pages 2440--2452, Marseille, France. European Language Resources
  Association.

\bibitem[{Ha et~al.(2019)Ha, Tran, {Ngo Van}, and Than}]{LDAdropout}
Cuong Ha, Van-Dang Tran, Linh {Ngo Van}, and Khoat Than. 2019.
\newblock \href {https://doi.org/10.1016/j.ijar.2019.05.010} {Eliminating
  overfitting of probabilistic topic models on short and noisy text: The role
  of dropout}.
\newblock \emph{International Journal of Approximate Reasoning}, 112:85--104.

\bibitem[{Hall et~al.(2008)Hall, Jurafsky, and
  Manning}]{hall-etal-2008-studying}
David Hall, Daniel Jurafsky, and Christopher~D. Manning. 2008.
\newblock \href {https://aclanthology.org/D08-1038} {Studying the history of
  ideas using topic models}.
\newblock In \emph{Proceedings of the 2008 Conference on Empirical Methods in
  Natural Language Processing}, pages 363--371, Honolulu, Hawaii. Association
  for Computational Linguistics.

\bibitem[{Hinton et~al.(2012)Hinton, Srivastava, Krizhevsky, Sutskever, and
  Salakhutdinov}]{HintonDropout}
Geoffrey~E. Hinton, Nitish Srivastava, Alex Krizhevsky, Ilya Sutskever, and
  Ruslan Salakhutdinov. 2012.
\newblock \href {http://arxiv.org/abs/1207.0580} {Improving neural networks by
  preventing co-adaptation of feature detectors}.
\newblock \emph{CoRR}, abs/1207.0580.

\bibitem[{Hoyle et~al.(2021)Hoyle, Goel, Hian-Cheong, Peskov, Boyd-Graber, and
  Resnik}]{hoyle2021is}
Alexander Hoyle, Pranav Goel, Andrew Hian-Cheong, Denis Peskov, Jordan~Lee
  Boyd-Graber, and Philip Resnik. 2021.
\newblock \href
  {https://proceedings.neurips.cc/paper/2021/file/0f83556a305d789b1d71815e8ea4f4b0-Paper.pdf}
  {Is automated topic model evaluation broken? the incoherence of coherence}.
\newblock In \emph{Advances in Neural Information Processing Systems}.

\bibitem[{Kingma and Welling(2014)}]{VAE}
Diederik~P. Kingma and Max Welling. 2014.
\newblock \href {http://arxiv.org/abs/1312.6114} {Auto-encoding variational
  bayes}.
\newblock In \emph{Proceedings of the 2nd International Conference on Learning
  Representations, {ICLR} 2014}.

\bibitem[{Labach et~al.(2019)Labach, Salehinejad, and Valaee}]{DropoutSurvey}
Alex Labach, Hojjat Salehinejad, and Shahrokh Valaee. 2019.
\newblock \href {http://arxiv.org/abs/1904.13310} {Survey of dropout methods
  for deep neural networks}.
\newblock \emph{CoRR}, abs/1904.13310.

\bibitem[{Lau et~al.(2014)Lau, Newman, and Baldwin}]{lau-etal-2014-machine}
Jey~Han Lau, David Newman, and Timothy Baldwin. 2014.
\newblock \href {https://doi.org/10.3115/v1/E14-1056} {Machine reading tea
  leaves: Automatically evaluating topic coherence and topic model quality}.
\newblock In \emph{Proceedings of the 14th Conference of the {E}uropean Chapter
  of the Association for Computational Linguistics}, pages 530--539,
  Gothenburg, Sweden. Association for Computational Linguistics.

\bibitem[{R\"{o}der et~al.(2015)R\"{o}der, Both, and Hinneburg}]{roder}
Michael R\"{o}der, Andreas Both, and Alexander Hinneburg. 2015.
\newblock \href {https://doi.org/10.1145/2684822.2685324} {Exploring the space
  of topic coherence measures}.
\newblock In \emph{Proceedings of the Eighth ACM International Conference on
  Web Search and Data Mining}, WSDM '15, page 399–408, New York, NY, USA.
  Association for Computing Machinery.

\bibitem[{Srivastava and Sutton(2017)}]{AVITM}
Akash Srivastava and Charles Sutton. 2017.
\newblock \href {https://openreview.net/forum?id=BybtVK9lg} {Autoencoding
  variational inference for topic models}.
\newblock In \emph{Proceedings of the 5th International Conference on Learning
  Representations, {ICLR} 2017}.

\bibitem[{Srivastava et~al.(2014)Srivastava, Hinton, Krizhevsky, Sutskever, and
  Salakhutdinov}]{SrivastavaDropout}
Nitish Srivastava, Geoffrey Hinton, Alex Krizhevsky, Ilya Sutskever, and Ruslan
  Salakhutdinov. 2014.
\newblock \href {http://jmlr.org/papers/v15/srivastava14a.html} {Dropout: A
  simple way to prevent neural networks from overfitting}.
\newblock \emph{Journal of Machine Learning Research}, 15(56):1929--1958.

\bibitem[{Terragni et~al.(2021)Terragni, Fersini, Galuzzi, Tropeano, and
  Candelieri}]{octis}
Silvia Terragni, Elisabetta Fersini, Bruno~Giovanni Galuzzi, Pietro Tropeano,
  and Antonio Candelieri. 2021.
\newblock \href {https://doi.org/10.18653/v1/2021.eacl-demos.31} {{OCTIS}:
  Comparing and optimizing topic models is simple!}
\newblock In \emph{Proceedings of the 16th Conference of the European Chapter
  of the Association for Computational Linguistics: System Demonstrations},
  pages 263--270. Association for Computational Linguistics.

\bibitem[{Tompson et~al.(2015)Tompson, Goroshin, Jain, LeCun, and
  Bregler}]{DropCNN}
Jonathan Tompson, Ross Goroshin, Arjun Jain, Yann LeCun, and Christoph Bregler.
  2015.
\newblock \href {https://doi.org/10.1109/CVPR.2015.7298664} {Efficient object
  localization using convolutional networks}.
\newblock In \emph{{IEEE} Conference on Computer Vision and Pattern
  Recognition, {CVPR} 2015, Boston, MA, USA, June 7-12, 2015}, pages 648--656.
  {IEEE} Computer Society.

\bibitem[{Warde{-}Farley et~al.(2014)Warde{-}Farley, Goodfellow, Courville, and
  Bengio}]{DropoutWarde}
David Warde{-}Farley, Ian~J. Goodfellow, Aaron~C. Courville, and Yoshua Bengio.
  2014.
\newblock \href {http://arxiv.org/abs/1312.6197} {An empirical analysis of
  dropout in piecewise linear networks}.
\newblock In \emph{2nd International Conference on Learning Representations,
  {ICLR} 2014}.

\bibitem[{Webber et~al.(2010)Webber, Moffat, and Zobel}]{Webber2010RBO}
William Webber, Alistair Moffat, and Justin Zobel. 2010.
\newblock \href {https://doi.org/10.1145/1852102.1852106} {A similarity measure
  for indefinite rankings}.
\newblock \emph{ACM Transactions on Information Systems (TOIS)}, 28(4):1--38.

\bibitem[{Wu and Gu(2015)}]{TowardsDropCNN}
Haibing Wu and Xiaodong Gu. 2015.
\newblock \href {https://doi.org/10.1016/j.neunet.2015.07.007} {Towards dropout
  training for convolutional neural networks}.
\newblock \emph{Neural Networks}, 71:1--10.

\bibitem[{Yan et~al.(2013)Yan, Guo, Lan, and Cheng}]{BTM}
Xiaohui Yan, Jiafeng Guo, Yanyan Lan, and Xueqi Cheng. 2013.
\newblock \href {https://doi.org/10.1145/2488388.2488514} {A biterm topic model
  for short texts}.
\newblock In \emph{Proceedings of the 22nd International Conference on World
  Wide Web}, WWW '13, page 1445–1456, New York, NY, USA. Association for
  Computing Machinery.

\bibitem[{Zhao et~al.(2021)Zhao, Phung, Huynh, Jin, Du, and
  Buntine}]{NTMsurvey}
He~Zhao, Dinh Phung, Viet Huynh, Yuan Jin, Lan Du, and Wray~L. Buntine. 2021.
\newblock \href {http://arxiv.org/abs/2103.00498} {Topic modelling meets deep
  neural networks: {A} survey}.
\newblock \emph{CoRR}, abs/2103.00498.

\bibitem[{Zhu et~al.(2018)Zhu, Feng, and Li}]{AllNews}
Qile Zhu, Zheng Feng, and Xiaolin Li. 2018.
\newblock \href {https://doi.org/10.18653/v1/D18-1495} {{G}raph{BTM}: Graph
  enhanced autoencoded variational inference for biterm topic model}.
\newblock In \emph{Proceedings of the 2018 Conference on Empirical Methods in
  Natural Language Processing}, pages 4663--4672, Brussels, Belgium.
  Association for Computational Linguistics.

\end{thebibliography}
\end{document}